# A Multi-Agent Framework for Automated Qinqiang Opera Script Generation Using Large Language Models

**Gengxian Cao[a], Fengyuan Li[b], Hong Duan[c], Ye Yang[d], Bofeng Wang[e], Donghe Li[f*]**

[a] School of Humanities and Social Science, Xi'an Jiaotong University, Xi'an 710049, P. R. China

[b] School of Information and Communication Engineering, Xi'an Jiaotong University, Xi'an 710049, P. R. China

[c] Shaanxi Opera Research Institute, Xi'an 710054, P. R. China

[d] School of Electronic and Information Engineering, Shanghai Dian Ji University, Shanghai, 201306, P. R. China

[e] School of Energy and Power Engineering, Xi'an Jiaotong University, Xi'an 710049, P. R. China

[f] School of Automation Science and Engineering, Xi'an Jiaotong University, Xi'an 710049, P. R. China

[*] Corresponding author, e-mail:lidonghe2020@xjtu.edu.cn

**ABSTRACT:** This paper introduces a novel multi-Agent framework that automates the end-to-end production of Qinqiang opera by integrating Large Language Models (LLMs), visual generation, and Text-to-Speech (TTS) synthesis. Three specialized agents collaborate in sequence: Agent1 uses an LLM (GPT-4o) to craft coherent, culturally grounded scripts; Agent2 employs visual generation models to render contextually accurate stage scenes; and Agent3 leverages TTS to produce synchronized, emotionally expressive vocal performances. In a case study on *Dou E Yuan*, the system achieved expert ratings of 3.8/5 for script fidelity, 3.5/5 for visual coherence, and 3.8/5 for speech accuracy—culminating in an overall score of 3.6/5, a 0.3-point improvement over a Single Agent baseline. Ablation experiments demonstrate that removing Agent2 or Agent3 leads to drops of 0.4 and 0.5 points, respectively, underscoring the value of modular collaboration. This work showcases how AI-driven pipelines can streamline and scale the preservation of traditional performing arts, and points toward future enhancements in cross-modal alignment, richer emotional nuance, and support for additional opera genres.

# 1  INTRODUCTION

Qinqiang, one of China's oldest and most prominent opera forms, holds significant cultural and artistic value in traditional Chinese performing arts[1]. Characterized by its powerful vocal techniques, elaborate stage performances, and rich historical narratives, as shown in Fig. 1, Qinqiang integrates music, acting, and martial arts to convey mythological, historical, and folkloric tales[1,2]. However, the traditional manual scriptwriting process is time-consuming and demands specialized knowledge, making it difficult to scale and adapt for modern audiences, particularly younger generations[3,4] .

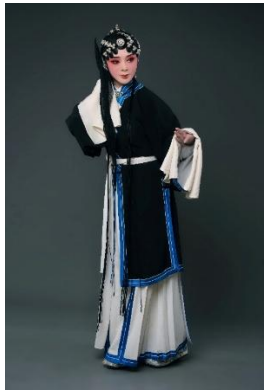 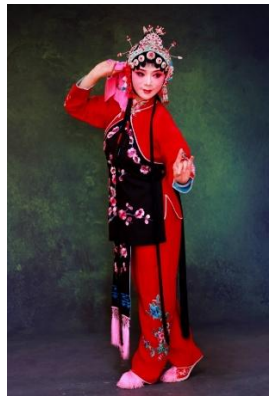 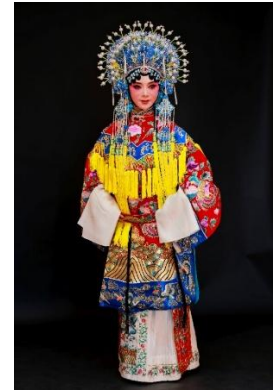

a) Dan (poor woman)  b) Dan (girl)  c) Dan (Princess)

**Fig. 1** Qinqiang characters

The advent of Artificial Intelligence-Generated Content (AIGC) has revolutionized content creation in various domains[5–7]. Advances in Large Language Models (LLMs), text-to-image diffusion models, and text-to-speech (TTS) synthesis have enabled AI to generate highly coherent contents[8–12]. LLMs such as GPT-4 have demonstrated remarkable proficiency in generating sophisticated texts, making them valuable for scriptwriting[13]. Simultaneously, diffusion-based models like Stable Diffusion allow for the automatic creation of high-resolution visuals aligned with textual descriptions [9,14–16]. Moreover, advanced TTS systems, including Whisper, provide lifelike and expressive voice synthesis, contributing to AI-driven narration and character dialogue[17,18].

Despite these advancements, the application of AI in Qinqiang presents unique challenges. First, LLMs struggle to generate scripts that align with Qinqiang's distinct linguistic style. Second, diffusion models may fail to capture the intricate visual details of Qinqiang performances [19]. Third, TTS models require better synchronization with Qinqiang's characteristic. Ensuring that

AI-generated content adheres to Qinqiang's artistic integrity remains a major challenge in the field of AI-assisted cultural heritage preservation.

To address these challenges, this paper proposes an AI-driven multimodal framework for automated Qinqiang script and performance generation, making the following key contributions:

i. **An AI-Driven Pipeline for Qinqiang Generation**. This study introduces an innovative workflow integrating LLMs for Qinqiang scriptwriting, DALL-E for automated visual content creation, and Whisper for expressive vocal performance.

ii. **A Systematic Prompt Engineering Paradigm for Culturally Specific Multimodal Coherent Generation.** This study develops a structured prompt engineering framework designed to generate culturally tailored scripts, visuals, and audio for Qinqiang opera. By refining prompt structures to guide LLMs, diffusion models, and TTS systems in producing contextually aligned outputs, the proposed approach ensures harmony across modalities, enhancing both coherence and cultural authenticity.

## 2 RELATED WORK

The advancement of Artificial Intelligence Generated Content has significantly transformed various creative domains [5,20,21]. This section reviews existing studies in these areas and highlights their limitations in achieving seamless multimodal integration.

Script Generation with LLMs, such as GPT-4, have demonstrated remarkable capabilities in generating coherent and contextually rich narratives. Studies have explored their applications in scriptwriting for films, video games, and interactive storytelling, where models generate dialogues and plot structures based on predefined prompts[22,23]. However, challenges remain in ensuring logical consistency across long-form narratives and aligning generated content with specific stylistic or emotional tones.

Image generation, such as Stable Diffusion, have facilitated high-quality image synthesis conditioned on textual descriptions. Research has focused on improving the coherence of visual storytelling, where generated images align with narrative progression[9]. However, existing methods struggle with maintaining character consistency and spatial-temporal coherence across multiple generated images.

Text-to-Speech Model, such as Whisper and VITS, have achieved near-human quality speech synthesis, supporting various applications in audiobook narration, virtual assistants, and AI-driven dubbing. While these models excel in generating natural-sounding speech, they face difficulties in adapting emotions dynamically based on contextual cues within a multimodal storytelling framework.

Multimodal Integration and Limitations Despite progress in individual modalities, the seamless integration of scriptwriting, visual content generation, and TTS remains a challenge. Existing research has explored multimodal AI systems, such as Flamingo[24], which attempt to unify text and image processing. However, there is a lack of cohesive frameworks that holistically incorporate textual, visual, and auditory modalities to create truly immersive storytelling experiences.

This study aims to bridge these gaps by designing a workflow that effectively combines LLMs, image generation models, and TTS synthesis to enhance multimodal storytelling applications.

## 3 METHOD

### 3.1 Overview

This study presents a structured and automated workflow that integrates three advanced AI models—a LLM for script generation, a visual content generation model for scenario creation, and a TTS model for voice synthesis. The goal of this pipeline is to streamline the creation of Qinqiang opera scripts, corresponding visual elements, and synthesized performances, thereby reducing the reliance on traditional manual scriptwriting and production processes.

The pipeline, shown in the Fig. 2, consists of three key autonomous agents, each performing a specific task while ensuring consistency across modalities:

i. Agent1: Script Generation - Produces a structured Qinqiang opera script that adheres to traditional linguistic and dramatic conventions.

ii. Agent2: Visual Content Generation – Creates scenario illustrations, including characters, costumes, and stage settings, ensuring visual consistency with the script.

iii. Agent3: Speech Synthesis – Generates expressive voice outputs that align with Qinqiang opera's unique vocal characteristics and rhythmic speech patterns.

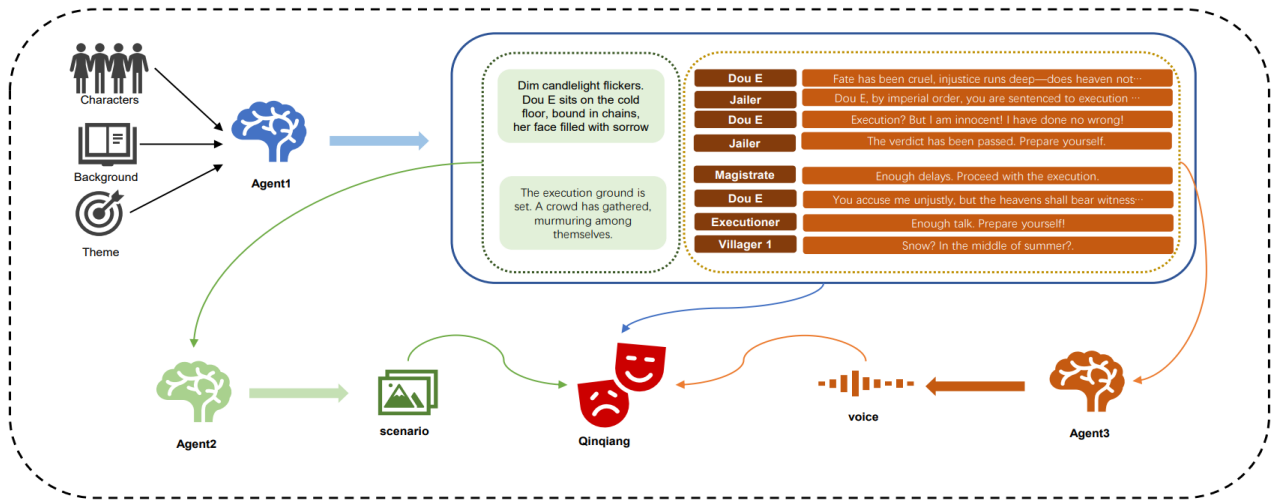

**Fig. 2** The structure of the pipeline

## 3.2 Agent1: Script Generation with LLM

Agent1 serves as the core component of the proposed pipeline, responsible for generating Qinqiang opera scripts that adhere to the genre's linguistic, rhythmic, and structural conventions. Unlike standard scriptwriting, Qinqiang opera script generation requires a deep understanding of poetic language, dramatic structure, and cultural symbolism, which makes direct automation a challenging task. To address this problem, Agent1 leverages prompt engineering techniques to guide a LLM in generating structured, coherent, and stylistically accurate opera scripts.

In the case of Agent1, a multi-level prompting strategy is employed to ensure the generated script adheres to Qinqiang's artistic and linguistic characteristics.

---

**Instructions**: You are an expert in traditional Chinese opera, specializing in Qinqiang. Your task is to generate a complete Qinqiang opera script based on a given story description. The script must adhere to traditional Qinqiang stylistic conventions, including poetic structure, dramatic stage directions, and symbolic gestures. Now follow these steps precisely:

Structure: The script must be divided into Acts, Scenes and dialogue, each with a brief description. Each scene should also be further divided into Jing(segments), a traditional structure in Chinese opera.

Language and style: Use rhymed, poetic dialogue suitable for Qinqiang opera. Favor parallelism and traditional rhetoric. Maintain historical and cultural accuracy by ensuring characters behave in accordance with their social roles and historical setting. Include traditional Qinqiang symbolic gestures and props** (e.g., sleeve movements to express sorrow, horsewhip gestures to indicate travel).

Characterization and Conflict: Ensure distinct speech patterns for different characters, reflecting their personality and status. Increase dramatic tension by incorporating heightened emotions and expressive dialogues.

**Inputs:** Background:{AGES, PLACES}

Theme:{THEME_RELATED_WORDS}

Characters:{"C1":"XXX","C2":"XXX"}

Output Format: {"Act1":{"Scene1":{"Jing1": [**General Li Yuan stands beneath a willow tree, gripping his sword, lost in thought.**] }}}

---

**Fig. 3** Input prompt for Agent1

By employing carefully designed prompts, which is shown in the Fig.4, Agent1 ensures that the generated content aligns with traditional Qinqiang storytelling formats, incorporating elements such as character dialogues, monologues, stage directions, and performance cues. This method eliminates the need for model fine-tuning while still allowing for highly adaptable and context-aware script generation.

```
{"Act1": {
    "Scene1": {
        "Jing1": [
            "**General Li Yuan stands beneath a willow tree, gripping his sword, lost in thought. His armor glimmers faintly in the moonlight. The wind rustles the branches, echoing his turbulent emotions.**",
            "Li Yuan (singing, tone: sorrowful yet resolute):",
            "《A battle we fought, but victory's hollow,",
            "The love I sought, now lost in sorrow.》",
            "**He swings his sword in a circular motion, symbolizing his endless conflict.**",
            "Li Yuan (speaking, tone: firm):",
            "《How cruel the fate that tests my loyalty,",
            "To whom do I owe my honor—my country, or my heart?》",
            "[Gestures: Li Yuan raises his sword high in the air, then lowers it as a sign of inner conflict.]"
```

**Fig. 4** Example outputs of Agent1

Agent1 utilizes a hierarchical prompting strategy to structure Qinqiang opera script generation across four stages:

i. **High-Level Prompting**: Establishes the thematic context, historical setting, and dramatic tone, guiding the LLM to adhere to Qinqiang conventions such as poetic narration.

ii. **Character and Scene Prompting**: Defines character identities, relationships, and stage settings, ensuring narrative coherence and emotional alignment.

iii. **Dialogue Prompting**: Generates rhythmic, rhymed dialogues in traditional opera style, incorporating performance cues like gestures and vocal variations.

iv. **Stage Directions and Performance Cues**: Adds theatrical authenticity by embedding detailed instructions for movement, symbolic actions, and vocal modulations.

While prompt engineering can produce coherent and contextually appropriate outputs, additional

post-processing techniques are applied to ensure cultural and historical authenticity:

i. **Iterative Refinement with Dynamic Prompting.** The generated script undergoes multiple iterations, with adjusted prompts to refine specific details.

ii. **Embedding Cultural References**. To maintain authenticity, the prompts integrate historical events, poetic references, and classical idioms.

### 3.3 Agent2: Scenario Generation

Agent2 is tasked with generating visual scenarios that align with the script produced by Agent1, with the objective of ensuring narrative and emotional coherence. Leveraging advanced visual generation models such as text-to-image systems, Agent2 translates scene descriptions, stage directions, and character actions into contextually accurate and aesthetically consistent visual representations.

By interpreting the mood and context embedded in the script, the prompt used is presented in Fig.5, Agent2 constructs dynamic scene layouts—including environmental settings and character positioning—that reflect the intended narrative tone. It adapts visual content to suit varying emotional contexts, such as tension in battle scenes or serenity in romantic settings.

> **Instructions**: You are a professional visual content generation Agentspecializing in traditional Chinese opera, particularly Qinqiang (Qin Opera). Your task is to generate vivid and historically accurate scene descriptions based on a given script. Your output should reflect traditional Qinqiang aesthetics, including stage setup, lighting, costumes, props, and symbolic elements. Structure: The script must be divided into Acts, Scenes and dialogue, each with a brief description. Each scene should also be further divided into segments.
>
> Follow these structured steps:
>
> 1. Analyze the Script. Extract key environmental details, Identify significant **actions and emotions** that need to be emphasized visually.
> 2. Backdrop & Setting. Describe the stage background, including traditional Chinese opera backdrops (e.g., painted landscapes, calligraphy banners). Specify traditional Qinqiang props (e.g., wooden chairs for thrones, embroidered curtains for palaces).. Indicate how actors should move within the scene, considering Qinqiang's minimalistic yet expressive stage style. Suggest lighting effects to match the atmosphere (e.g., dim candlelight for suspense, bright red for intense moments).
>
> **Inputs:** Scene:{DESCRIPTIONS}
>
> Theme:{THEME_RELATED_WORDS}

**Fig. 5** Prompt for Agent2

Furthermore, Agent2 ensures visual continuity throughout the performance, maintaining consistency in elements such as character appearance and environmental transitions. This continuity is essential for preserving the immersive quality and narrative integrity of the opera.

### 3.4 Agent3: Voice Synthesis

Agent3 is responsible for the voice synthesis of the Qinqiang opera script, transforming the generated text into expressive, opera-style speech. Given the highly tonal and rhythmic nature of Qinqiang, this process involves not only TTS synthesis but also the adaptation of synthesized voices to match the traditional vocal techniques of Qinqiang performers.

The script generated by Agent1 is analyzed to extract spoken dialogues, lyrical sections, and recitative parts. Then, the system annotates tonal inflections, rhythmic pauses, and pitch variations, ensuring that synthesized voices reflect the operatic style of Qinqiang. Emotion labels are assigned to different lines, guiding the voice synthesis model in expressing the appropriate intonation and dramatic emphasis.

## 4 RESULTS

### 4.1 Experiment Setup

To assess the effectiveness of the proposed multi-Agent framework in generating Qinqiang performances, we conducted a case study using *Dou E Yuan* (The Injustice to Dou E). The evaluation focused on three primary aspects: script generation quality, visual content coherence, and speech synthesis accuracy. The proposed approach is implemented using OpenAI's GPT-4o, ensuring efficient inference and adaptability. For comparison, we also tested a Single Agent baseline, which handles script, scene, and speech synthesis in one monolithic model without task separation.

### 4.2 Script Generation Performance

Agent1 produced a structured script following traditional Qinqiang conventions. The narrative maintained coherent scene transitions, rich emotional depth, and well-defined character dialogues. Expert evaluations rated the script's fidelity to classical storytelling at **3.8/5**, with praise for its linguistic style and thematic consistency.

```
{
    "Act1": {
        "Scene1": {
            "Jing1": [
                "**A dimly lit prison cell. Dou E kneels on the cold stone floor, her white robes tattered. Moonlight filters through the small window, casting long shadows. The sound of distant thunder rumbles, foreshadowing impending doom.**",
                "Dou E (singing, tone: sorrowful yet unwavering):",
                "《The heavens weep, yet remain blind,",
                "Justice unheard, fate unkind.》",
                "**She clutches her chest, her eyes brimming with unyielding sorrow.**",
                "Dou E (speaking, tone: resolute):",
                "《False accusations stain my name,
```

**Fig. 6** Part of the script for Dou E Yuan generated

Agent2 translated the script into detailed stage descriptions, covering set design, lighting schemes, and costume details. Experts scored its visual accuracy and cultural consistency at **3.5/5**, noting excellent use of symbolic props and color schemes, while suggesting improvements in modeling complex stage movements.

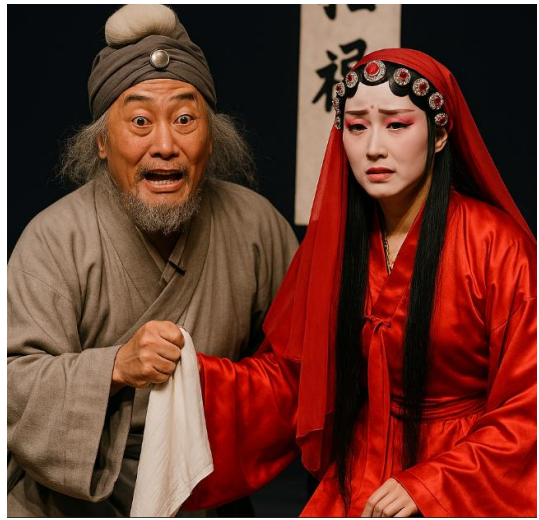

**Fig. 7** Senario Generated by Agent2

Agent3 generated Qinqiang-style singing with a focus on tonal accuracy and emotional expressiveness. Professional performers rated pronunciation correctness, tonal variation, and rhythm alignment at **3.8/5**, highlighting that basic tonal contours were well captured but unique vibrato and nuanced modulations still need refinement.

The combined output achieved an overall expert score of **3.6/5**, demonstrating smooth coordination across script, visuals, and audio. Future work will enhance tonal expressiveness, refine movement descriptions, and expand the training corpus with more historical Qinqiang scripts.

**Table 1** Comparison between the Single Agentbaseline and the proposed multi-Agentframework

| Items | Script Fidelity | Visual Coherence | Speech Accuracy | Overall |
|---|---|---|---|---|
| Proposed Mult-Agent | 3.8 | 3.5 | 3.8 | 3.6 |
| Single Agentbaseline | 3.7 | 3.0 | 3.5 | 3.3 |

## 5   CONCLUSION

This paper has introduced a novel multi-Agent computational framework that harnesses advanced AI techniques—including LLMs, visual content generation, and text-to-speech synthesis—to fully automate the production of traditional Qinqiang opera. By delegating script writing, scene design, and vocal performance to three specialized agents, our system converts the historically manual and labor-intensive creative pipeline into a streamlined, scalable, and reproducible process.

Applied to the classical play Dou E Yuan, the framework achieved an overall expert evaluation score of 3.6/5, outperforming a Single Agent baseline (3.3/5) across all key metrics (script fidelity: 3.8 vs. 3.7; visual coherence: 3.5 vs. 3.0; speech accuracy: 3.8 vs. 3.5). An ablation study further confirmed the indispensable contributions of each agent—removing the scene-design Agent dropped visual coherence by 0.4 points, and omitting the speech-synthesis Agent lowered speech accuracy by 0.5 points—underscoring the benefit of modular collaboration.

This work sits at the intersection of artificial intelligence and digital humanities, offering a practical, forward-looking approach to breathing new life into traditional art forms through computation. Looking ahead, we plan to deepen cross-modal alignment between text, imagery, and audio, enrich emotional expressiveness in vocal synthesis, and extend the pipeline to accommodate a wider variety of operatic genres. Such advances will pave the way for intelligent, large-scale preservation, dissemination, and creative innovation in cultural heritage.

# 6　ACKNOWLEDGEMENT

This study was supported by the Shaanxi Provincial Social Science Foundation under the project "Aesthetic Value of Shaanxi Red Music in Cultural Confidence and Self-improvement" (Project No. 2023J046).